\documentclass[sigconf]{acmart}

\usepackage{bm} 
\usepackage{multirow} 
\usepackage{booktabs} 
\usepackage{color}
\definecolor{ggray}{RGB}{127,127,127}

\usepackage{cleveref}
\crefname{section}{§}{§§}
\Crefname{section}{§}{§§}
\newcommand{\eg}{\textit{e}.\textit{g}.}  
\newcommand{\ie}{\textit{i}.\textit{e}.}  
\newcommand{\tabincell}[2]{\begin{tabular}{@{}#1@{}}#2\end{tabular}} 
\newcommand{\etal}{\textit{et al.}}  
\usepackage{url}  
\usepackage{balance}

\copyrightyear{2021}
\acmYear{2021}
\setcopyright{acmcopyright}\acmConference[MM '21]{Proceedings of the 29th ACM International Conference on Multimedia}{October 20--24, 2021}{Virtual Event, China}
\acmBooktitle{Proceedings of the 29th ACM International Conference on Multimedia (MM '21), October 20--24, 2021, Virtual Event, China}
\acmPrice{15.00}
\acmDOI{10.1145/3474085.3475485}
\acmISBN{978-1-4503-8651-7/21/10}

\begin{document}

\title{Improving Weakly-supervised Object Localization \\ via Causal Intervention}



\author{Feifei Shao}
\affiliation{%
  \institution{Zhejiang University}
  \city{Hangzhou}
  \country{China}}
\email{sff@zju.edu.cn}

\author{Yawei Luo}
\authornote{Yawei Luo is the corresponding author}
\affiliation{%
  \institution{Zhejiang University}
  \city{Hangzhou}
  \country{China}}
\affiliation{%
  \institution{Baidu Research}
  \city{Beijing}
  \country{China}
}
\email{yaweiluo329@gmail.com}

\author{Li Zhang}
\affiliation{%
  \institution{Zhejiang Insigma Digital Technology Co., Ltd.}
  \city{Hangzhou}
  \country{China}}
\email{zhangli@insigma.com.cn}

\author{Lu Ye}
\affiliation{%
  \institution{Zhejiang University of Science and Technology}
  \city{Hangzhou}
  \country{China}}
\email{yelue@zust.edu.cn}

\author{Siliang Tang}
\affiliation{%
  \institution{Zhejiang University}
  \city{Hangzhou}
  \country{China}}
\email{siliang@zju.edu.cn}

\author{Yi Yang}
\affiliation{%
  \institution{Zhejiang University}  
  \city{Hangzhou}
  \country{China}}
\email{yee.i.yang@gmail.com}

\author{Jun Xiao}
\affiliation{%
  \institution{Zhejiang University}
  \city{Hangzhou}
  \country{China}}
\email{junx@cs.zju.edu.cn}








\renewcommand{\shortauthors}{Shao and Luo, et al.}

\begin{abstract}
  
  The recently emerged weakly-supervised object localization (WSOL) methods can learn to localize an object in the image only using image-level labels. Previous works endeavor to perceive the interval objects from the small and sparse discriminative attention map, yet ignoring the co-occurrence confounder (\eg, duck and water), which makes the model inspection (\eg, CAM) hard to distinguish between the object and context. In this paper, we make an early attempt to tackle this challenge via causal intervention (CI). Our proposed method, dubbed CI-CAM, explores the causalities among image features, contexts, and categories to eliminate the biased object-context entanglement in the class activation maps thus improving the accuracy of object localization. Extensive experiments on several benchmarks demonstrate the effectiveness of CI-CAM in learning the clear object boundary from confounding contexts. Particularly, on the CUB-200-2011 which severely suffers from the co-occurrence confounder, CI-CAM significantly outperforms the traditional CAM-based baseline ($58.39\%$ \emph{vs} $52.4\%$ in Top-1 localization accuracy). While in more general scenarios such as ILSVRC 2016, CI-CAM can also perform on par with the state of the arts.

\end{abstract}

\ccsdesc[500]{Computing methodologies~Interest point and salient region detections}
\ccsdesc[500]{Computing methodologies~Object detection}
\ccsdesc[500]{Computing methodologies~Object recognition}

\keywords{Object Localization; Causal Intervention; Weakly-supervised Learning}
\maketitle
\section{Introduction}

Object localization~\cite{tompson2015efficient, choudhuri2018object} aims to indicate the category, the spatial location and scope of an object in a given image, in forms of bounding box~\cite{everingham2010pascal, russakovsky2015imagenet}. This task has been studied extensively in the computer vision community~\cite{tompson2015efficient} due to its broad applications, such as scene understanding and autonomous driving. Recently, the techniques based on deep convolutional neural networks (DCNNs)~\cite{simonyan2014very,szegedy2015going,he2016deep,luo2021category,luo2018macro} promote the localization performance to a new level. However, this performance promotion is at the price of huge amounts of fine-grained human annotations\cite{luo2019taking,luo2020ASM}. To alleviate such a heavy burden, weakly-supervised object localization (WSOL) has been proposed by only resorting to image-level labels. 

\begin{figure}[t]
  \centering
  \includegraphics[width=0.9\linewidth]{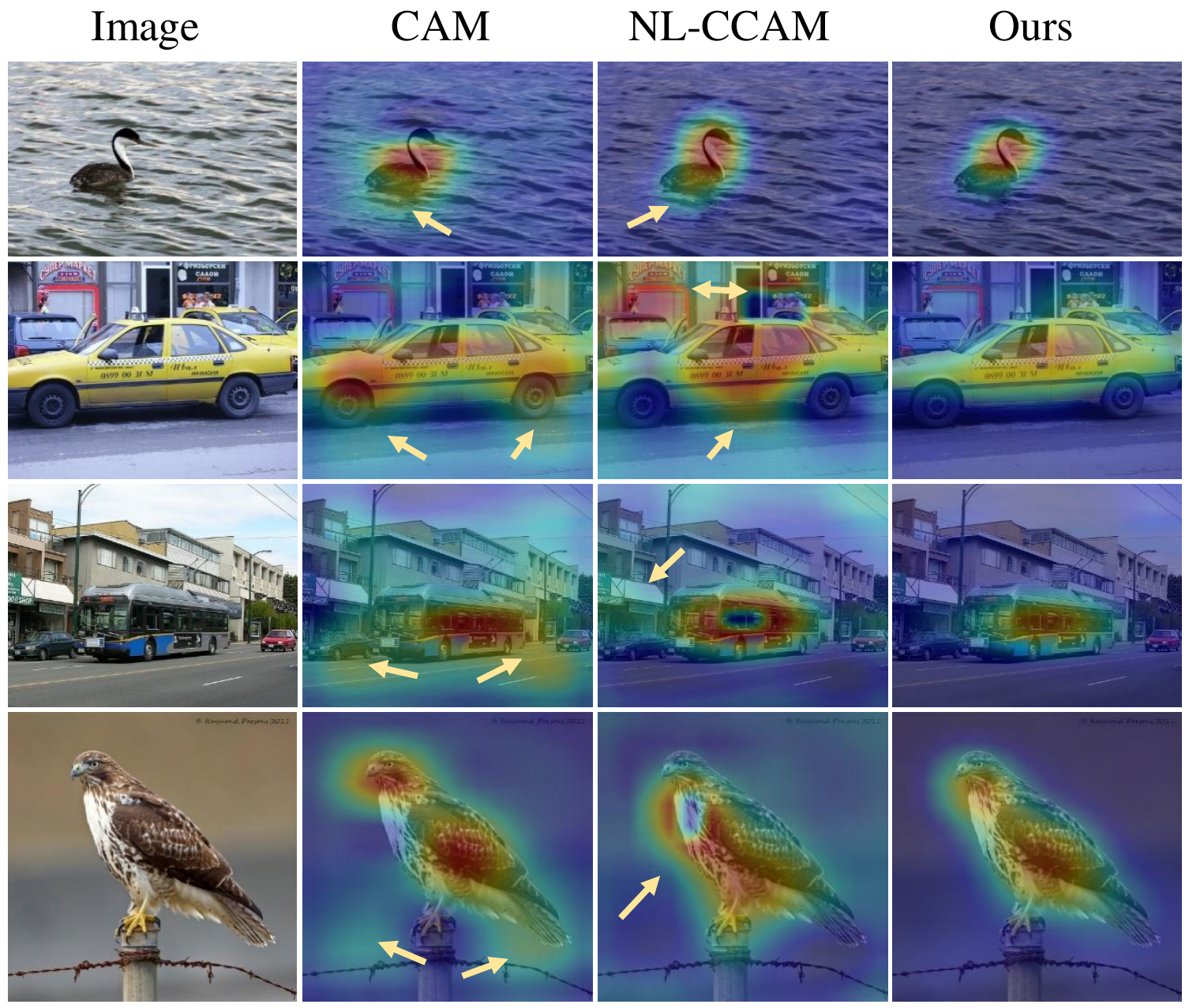}
  \caption{Visualization comparison between CAM, NL-CCAM and CI-CAM (ours). The yellow arrows indicate the regions suffer from entangled context.}
  \label{comparison_graph}
\end{figure}

To capitalize on the image-level labels, the existing studies~\cite{selvaraju2017grad, diba2017weakly, kim2017two, wei2018ts2c, shao2021deep} follow the Class Activation Mapping (CAM) approach~\cite{zhou2016learning} to generate class activation maps first and segment the highest activation area for a coarse localization. Albeit, CAM is initially designed for the classification task and tends to focus only on the most discriminative feature to increase its classification accuracy. Targeting on this issue, recent prevailing works~\cite{diba2017weakly, kim2017two, wei2018ts2c, zhang2018adversarial, gao2019c, mai2020erasing, yang2020combinational} endeavor to perceive the interval objects from the small and sparse ``discriminative regions''. On one hand, they adjust the network structure to make the detector more tailored for object localization in a weak supervision setting. For example, some methods~\cite{diba2017weakly, wei2018ts2c} use a three stages network structure to continuously optimize the prediction results by training the current stage using the output of the previous stages as supervision. Besides, some methods~\cite{kim2017two, zhang2018adversarial, gao2019c, mai2020erasing} use two parallel branches that its first branch is responsible for digging out the most discriminative small regions, and the second branch is responsible for detecting the second discriminative large regions. On the other hand, they also make full use of the image information to improve the prediction results. For example, $TS^2C$~\cite{wei2018ts2c} and NL-CCAM~\cite{yang2020combinational} utilize the contextual information of surrounding pixels and the activation maps of low probability class, respectively.


In contrast to the prevailing efforts focusing on the most discriminative feature, in this work we target another key issue which we called ``entangled context''. The reason behind this issue is that the objects usually co-occur with a certain background. For example, if most ``duck'' appears concurrently with ``water'' in the images, these two concepts would be inevitably entangled and a classification model would wrongly generate ambiguous boundary between ``duck'' and ``water'' with only image-level supervision. In contrast to the vanilla CAM which yields a relatively small bounding box on the small discriminative region, we notice that our targeted problem would cause a biased bounding box that includes the wrongly entangled background, which impairs the localization accuracy in terms of the object range. Consequently, we argue that resolving the ``entangled context'' issue is vital for WSOL but remains unnoticed and unexplored.

In this paper, we propose a principled solution, dubbed CI-CAM, tailored for WSOL based on Causal Inference~\cite{pearl2014interpretation} (CI). CI-CAM ascribes the ``entangled context'' issue to the frequently co-occurred background that misleads the image-level classification model to learn spurious correlations between pixels and labels. To find those intrinsic pixels which truly cause the image-level label, CI-CAM first establishes a structural causal model (SCM)~\cite{pearl2016causal}. Based on SCM, CI-CAM further regards the object context as a confounder and explores the causalities among image features, contexts, and labels to eliminate the biased co-occurrence in the class activation maps. More specifically, the CI-CAM causal context pool accumulates the contextual information of each image for each category and employs it as attention in convolutional layers to enhance the feature representation for making feature boundary clearer. To our knowledge, we are making an early attempt to apprehend and approach the ``entangled context'' issue for WSOL. To sum up, the contributions of this paper are as follows.

\begin{itemize}
   \item We are among the pioneers to concern and reveal the ``entangled context'' issue of WSOL that remains unexplored by prevailing efforts.
   \item We propose a principled solution for the ``entangled context'' issue based on causal inference, in which we pinpoint the context as a confounder and eliminate its negative effects on image features via backdoor adjustment~\cite{pearl2016causal}.
   \item We design a new network structure, dubbed CI-CAM, to embed the causal inference into the WSOL pipeline with an end-to-end scheme.
   \item  The proposed CI-CAM achieves state-of-the-art performance on the CUB-200-2011 dataset~\cite{wah2011caltech}, which significantly outperforms the baseline method by $2.16\%$ and $5.99\%$ in terms of classification and localization accuracy, respectively. While in more general scenarios such as ILSVRC 2016~\cite{russakovsky2015imagenet} which suffer less from the ``entangled context'' due to the huge amount of images and various backgrounds, CI-CAM can also perform on par with the state of the arts.
\end{itemize}

\section{Related Work}
\subsection{Weakly-supervised Object Localization}

Since CAM~\cite{zhou2016learning} is prone to bias to the most discriminative part of the object rather than the integral object, the research attention of most of the current methods is how to improve the accuracy of object localization. These methods can be broadly categorized into two groups: enlarging the proposal region and discriminative region removal. 

1) \textbf{Enlarging proposal region}: enlarging the box size appropriately of the initial prediction box~\cite{diba2017weakly, wei2018ts2c}. WCCN~\cite{diba2017weakly} introduces three cascaded networks trained in an end-to-end pipeline. The latter stage continuously enlarges and refines the output proposals of its previous stage. $TS^2C$~\cite{wei2018ts2c} selects the box by comparing the mean pixel confidence values of the initial prediction region and its surrounding region. If the gap of the mean values of two regions is large, the initial prediction region is the final prediction box; otherwise, the surrounding region. 

2) \textbf{Discriminative region removal}: detecting the bigger region after removing the most discriminative region~\cite{kim2017two, zhang2018adversarial, choe2019attention, mai2020erasing}. TP-WSL~\cite{kim2017two} first detects the most discriminative region in the first network, Then, it erases this region of the conv5-3 feature maps in the second network (\eg, zero). ACoL~\cite{zhang2018adversarial} uses the masked feature maps by erasing the most discriminative region discovered in the first classifier as the input feature maps of the second classifier. ADL~\cite{choe2019attention} stochastically produces an erased mask or an importance map at each iteration as a final attention map projected in the feature maps of images. EIL~\cite{mai2020erasing} is an adversarial erasing method that simultaneously computes the erased branch and the unerased branch by sharing one classifier. 


The above methods basically focus on the poor localization caused by the most discriminative part of the object. However, they ignore the problem of the fuzzy boundary between the objects and the co-occur certain context background. For example, if most ``duck'' appears concurrently with ``water'' in the images, these two concepts would be inevitably entangled and wrongly generate ambiguous boundaries using only image-level supervision.


\subsection{Causal Inference}
Causal inference~\cite{sobel1996introduction, yao2020survey, pearl2009causal} is a critical research topic across many domains, such as statistics~\cite{pearl2016causal}, politics~\cite{keele2015statistics}, psychology, and epidemiology~\cite{mackinnon2007mediation, richiardi2013mediation}. The purpose of causal inference is to give the model the ability to pursue causal effects: we can eliminate false bias~\cite{bareinboim2012controlling}, clarify the expected model effects~\cite{besserve2018counterfactuals}, and modularize reusable features to make them well generalized~\cite{parascandolo2018learning}. Nowadays, causal inference is used repeatedly in computer vision tasks~\cite{niu2020counterfactual, chen2020counterfactual, qi2020two, tang2020long, tang2020unbiased, wang2020visual, yang2020deconfounded, yue2020interventional, zhang2020causal}. Specifically, Zhang \etal~\cite{zhang2020causal} utilize a SCM~\cite{didelez2001judea, pearl2016causal} to deeply analyze the
causalities among image features, contexts, and class labels and propose a new network: Context Adjustment (CONTA) that achieves the new state-of-the-art performance in weakly-supervised semantic segmentation task. Yue \etal~\cite{yue2020interventional} use the causal intervention in few-shot learning. They uncover the pre-trained knowledge is indeed a confounder that limits the performance. And they propose a novel FSL paradigm: Interventional Few-Shot Learning (IFSL), which is implemented via the backdoor adjustment~\cite{pearl2016causal}. Tang \etal~\cite{tang2020long} show that the SGD momentum is essentially a confounder in long-tailed classification by using a SCM.

In our work, we also leverage a SCM~\cite{pearl2016causal} to analyze the causalities among image features, contexts, and class labels, we find that context is a confounder factor. In \S\ref{sec:network_structure}, we will introduce a causal context pool that is used for eliminating the negative effects of the context and keeping its positive effects.


\section{Methodology}
\begin{figure*}[t]
   \centering
   \includegraphics[width=0.9\linewidth]{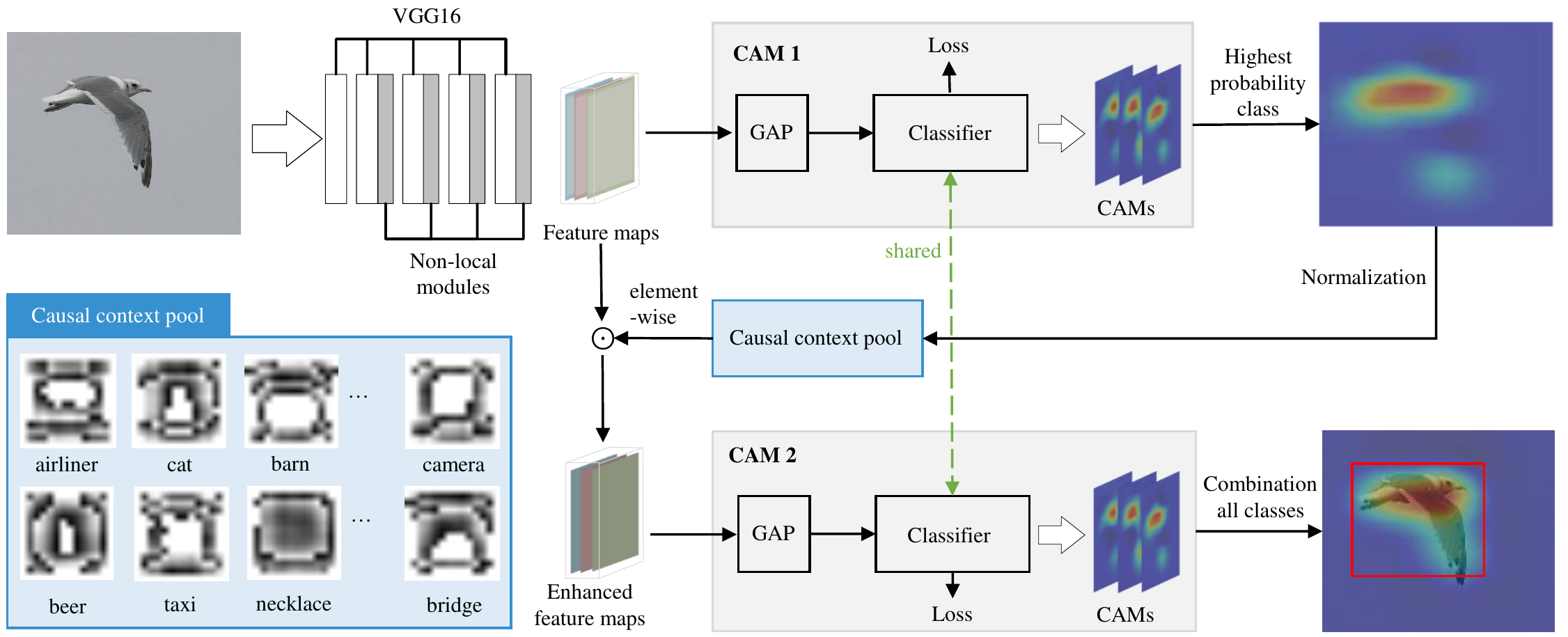}
   \caption{Overview of the proposed CI-CAM approach. CI-CAM consists of four parts: a backbone to extract the feature maps, the share-weighted CAM modules to generate the class activation maps, a causal context pool (which is the core of the CI-CAM method) to enhance the feature maps by eliminating the negative effect of confounder, and a combinational module to generate the final bounding box.}
   \label{network_architecture_graph}
 \end{figure*}

In this section, we describe the details of the CI-CAM method as shown in Figure~\ref{network_architecture_graph}. We first introduce the preliminaries of CI-CAM including problem settings, causal inference, and baseline method in \S\ref{sec:preliminaries}. Second, we formulate the causalities among pixels, context, and labels with a SCM in \S\ref{sec:structural_causal_model}. Based on SCM, we approach the ``entangled context'' issue in a principled way via causal inference in \S\ref{sec:causal intervention for WSOL}. We design the network structure of CI-CAM to embed causal inference in the WSOL pipeline detailed in \S\ref{sec:network_structure}, at the core of which is the causal context pool. Finally, we give the training objective of CI-CAM in \S\ref{sec:training_objective}.

\subsection{Preliminaries}
\label{sec:preliminaries}
\textbf{Problem Settings.} Before presenting our method, we first introduce the problem settings of WSOL formally. Given an image $I$, WSOL targeting at classifying and locating one object in terms of the class label and the bounding box. However, only image-level labels $Y$ can be accessed during the training phase.

\textbf{Causal Inference.} Causal inference enables to equip the model with the ability to pursue causal effects: it can eliminate false bias~\cite{bareinboim2012controlling} as well as clarify the expected model effects~\cite{besserve2018counterfactuals}. SCM~\cite{pearl2016causal} is a directed graph in which each node represents each participant of the model, and each link denotes the causalities between the two nodes. Nowdays, SCM is widely used in causal inference scenes~\cite{yue2020interventional, zhang2020causal, tang2020long}. Backdoor adjustment~\cite{pearl2016causal} is responsible for finding the wrong impact between two nodes and eliminating this issue by leveraging the three do-calculus rules~\cite{neuberg2003causality}.

\begin{figure}[t]
   \centering
   \includegraphics[width=0.9\linewidth]{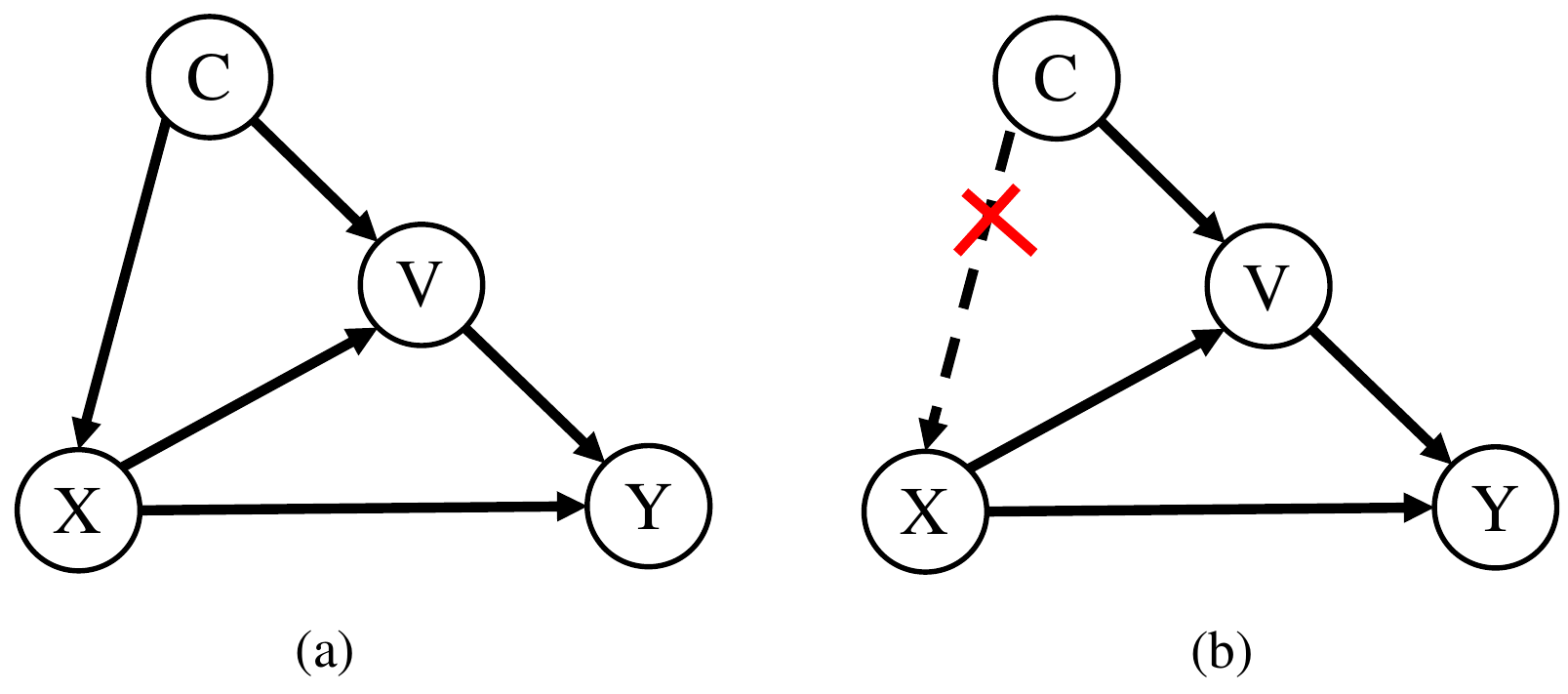}
   \caption{(a) The structural causal model (SCM) for causality of classifier in WSOL, (b) The intervened SCM for the causality of classifier in WSOL.}
   \label{structural_causal_model_graph}
 \end{figure}

\textbf{Baseline Method.}
Class activation maps (CAMs) are widely employed for locating the object boxes in the WSOL task. Yang \etal~\cite{yang2020combinational} argue that using only one activation map of the highest probability class for segmenting object boxes is problematic since it often biases into limited regions or sometimes even highlights background regions. Based on such observation, they propose the NL-CCAM~\cite{yang2020combinational} method to combine all activation maps from the highest to the lowest probability class to a localization map using a specific combinational function and achieves good localization performance.

Based on vanilla fully convolutional network (FCN)-based backbone, \eg, VGG16~\cite{simonyan2014very}, NL-CCAM~\cite{yang2020combinational} inserts four non-local blocks before every bottleneck layer excluding the first bottleneck layer simultaneously to produce a non-local fully convolutional network (NL-FCN). Given an image $I$, it is fed into the NL-FCN to produce its feature maps $X \in \mathbb{R}^{c \times h \times w}$, where $c$ is the number of channels and ${h \times w}$ is the spatial size. Then, the feature maps $X$ are forwarded to a global average pooling (GAP) layer followed by a classifier with a fully connected layer. The prediction scores $S=\{s_1, s_2, \ldots, s_n\}$ are computed by using a softmax layer on the top of the classifier for classification. The weight matrix of the classifier is denoted as $W \in \mathbb{R}^{n \times c}$, where $n$ is the number of image classes. Therefore, the activation maps $M_i$ of class $i$ among class activation maps (CAMs) $M \in \mathbb{R}^{n \times h \times w}$ proposed in~\cite{zhou2016learning} are given as follows.

\begin{equation}
   \begin{aligned}
      M_i = \sum_{k}^{c} {W_{i, k} \cdot X_k},
   \end{aligned}
   \label{eq:cam}
\end{equation}
where $i \in \{1,2, \ldots, n\}$.

NL-CCAM~\cite{yang2020combinational} produces a localization map by using a combinational function in CAMs instead of using the activation map of the highest probability class among CAMs. Firstly, it ranks the activation maps from the highest probability class to the lowest and uses $M_{t_k}$ to denote the activation map of the $k$ highest probability class. The class label with the highest probability $t_1$ is computed as follows.
\begin{equation}
   \begin{aligned}
      t_1 = argmax_k(\{S_k\}),
   \end{aligned}
\end{equation}
where $k \in \{1,2, \ldots, n\}$. Then it combines the class activation maps $M$ to a localization map $H \in \mathbb{R}^{h \times w}$ as follows.
\begin{equation}
   \begin{aligned}
      H = \sum_{k}^n \gamma(k)M_k,
   \end{aligned}
   \label{eq:ccam}
\end{equation}
where $\gamma(\cdot)$ is a combinational function. Finally, it segments the localization map $H$ using a thresholding technique proposed in~\cite{zhou2016learning} to generate a bounding box for object localization.

Our method is based on NL-CCAM~\cite{yang2020combinational} but introduces substantial improvements. We equip the baseline network with the ability of causal inference to tackle the ``entangled context'' issue, which will be detailed in the following.

\subsection{Structural Causal Model}
\label{sec:structural_causal_model}

In this section, we will reveal the reason why the context hurts the object localization quality. We formulate the causalities among image features $X$, context confounder $C$, and image-level labels $Y$, with a structural causal model (SCM)~\cite{pearl2016causal} shown in Figure~\ref{structural_causal_model_graph} (a). The direct links denote the causalities between the two nodes: cause $\rightarrow$ effect.

$\bm{C \rightarrow X}$: This link indicates that the backbone generates feature maps $X$ under the effect of context $C$. Although the confounder context $C$ is helpful for a better association between the image features $X$ and labels $Y$ via a model $P(Y\mid X)$, \textit{e.g.}, it is likely a ``duck'' when seeing a ``water'' region, $P(Y\mid X)$ mistakenly associates non-causal but positively correlated pixels to labels, \textit{e.g.}, the ``water'' region wrongly belongs to ``duck''. That is one reason for the inaccurate localization in WSOL. Fortunately, as we will introduce later in \S\ref{sec:network_structure}, we can avoid it by using a causal context pool in causal intervention. 

$\bm{C \rightarrow V \leftarrow X}$: $V$ is an image specific-representation using the contextual templates from $C$. For example, $V$ tells us that the shape and location of a ``duck'' (foreground) in a scene (background). Note that this assumption is not adhoc in our model, it underpins almost every concept learning method from the classic Deformable Part Models~\cite{felzenszwalb2009object} to modern CNNs~\cite{girshick2015deformable}.

$\bm{X \rightarrow Y \leftarrow V}$: These links indicate that $X$ and $V$ together affect the label $Y$ of an image. $V \rightarrow Y$ denotes that the contextual templates directly affect the image label. It is worth noting that even if we do not explicitly take $V$ as an input for the WSOL model, $V \rightarrow Y$ still holds. On the contrary, if $V \nrightarrow Y$ in Figure~\ref{structural_causal_model_graph} (a), the only path left from $C$ to $Y$: $C \rightarrow X \rightarrow Y$. Considering the effect of context $C$ on image features $X$, we will cut off the link of $C$ between $X$. Then, no context $C$ are allowed to contribute to the image label $Y$ by training $P(Y\mid X)$, which results in never uncovering the context. So, WSOL would be impossible.

So far, we have pinpointed the role of context $C$ played in the causal graph of image-level classification in Figure~\ref{structural_causal_model_graph} (a). Next, we will introduce a causal intervention method to remove the confounding effect.

\subsection{Causal Intervention for WSOL}
\label{sec:causal intervention for WSOL}
We propose to use $P(Y\mid do(X))$ based on the backdoor adjustment~\cite{pearl2016causal} as the new image-level classifier, which removes the confounder $C$ and pursues the true causality from $X$ to $Y$ shown in Figure~\ref{structural_causal_model_graph} (b). By this way, we can achieve better classification and localization in WSOL. The key idea is to 1) cut off the link $C \rightarrow X$, and 2) stratify $C$ into pieces $C = \{c_1, c_2, \ldots, c_n\} $, and $c_i$ denotes the $i^{th}$ class context. Formally, we have 
\begin{equation}
   \begin{aligned}
      P(Y\mid do(X)) = \sum_i^n P(Y\mid X=x, V=f(x, c_i))P(c_i),
   \end{aligned}
   \label{eq:p_do_x}
\end{equation}
where $f(X,c)$ abstractly represents that $V$ is formed by the combination of $X$ and $c$, and $n$ is the number of image class. As $C$ does not affect $X$, it guarantees $X$ to have a fair opportunity to incorporate every context $c$ into $Y$'s prediction, subject to a prior $P(c)$. To simplify the forward propagation of the network, we adopt the Normalized Weighted Geometric Mean~\cite{xu2015show} to optimize Eq.~\eqref{eq:p_do_x} by moving the outer sum $\sum_i^n P(c_i)$ into the feature level
\begin{equation}
   \begin{aligned}
      P(Y\mid do(X)) \approx P(Y\mid X=x, V=\sum_i^n f(x, c_i)P(c_i)).
   \end{aligned}
   \label{eq:p_do_x_nwgm}
\end{equation}
Therefore, we only need to feed-forward the network once instead of $n$ times. Since the number of samples for each class in the dataset is roughly the same, we set $P(c)$ to uniform $1/n$. After further optimizing Eq.~\eqref{eq:p_do_x_nwgm}, we have
\begin{equation}
   \begin{aligned}
      P(Y\mid do(X)) \approx P(Y\mid x \oplus \frac{1}{n}\sum_i^n f(x, c_i)),
   \end{aligned}
   \label{eq:p_do_x_final}
\end{equation}
where $\oplus$ denotes projection. So far, the ``entangled context'' issue has been transferred into calculating $\sum_i^n f(x, c_i)$. We will introduce a causal context pool $Q$ to represent $\sum_i^n f(x, c_i)$ in \S\ref{sec:network_structure}.

\subsection{Network Structure}
\label{sec:network_structure}
In this section, we implement causal inference for WSOL with a tailored network structure, at the core of which is a causal context pool. The main idea of the causal context pool is to accumulate all contexts of each class, and then re-project the contexts to the feature maps of convolutional layers shown in Eq.~\eqref{eq:p_do_x_final} to pursue the pure causality between the cause $X$ and the effect $Y$. Figure~\ref{network_architecture_graph} illustrates the overview of CI-CAM that includes four parts: backbone, CAM module, causal context pool, and combinational part.

\textbf{Backbone.} Inherited from the baseline method, we design our backbone by inserting multiple non-local blocks at both low- and high-level layers of a FCN-based network simultaneously. It acts as a feature extractor that takes the RGB images as input and producing high-level position-aware feature maps.

\textbf{CAM module.} It includes a global average pooling (GAP) layer and a classifier with a fully connected layer~\cite{zhou2016learning}. Image feature maps $X$ generated by the backbone are fed into GAP and classifier to produce prediction scores $S=\{s_1, s_2, \ldots, s_n\}$. Then, the CAM network multiplies the weight $W$ of the classifier to $X$ to produce class activation maps $M \in \mathbb{R}^{n \times h \times w}$ shown in Eq.~\eqref{eq:cam}. In our model, we use two CAM modules with shared weights. The first CAM module is designed to produce initial prediction scores $S$ and class activation maps $M$, and the second CAM network is responsible for producing more accurate prediction scores $S^e=\{s_1^e, s_2^e, \ldots, s_n^e\}$ and class activation maps $M^e \in \mathbb{R}^{n \times h \times w}$ using the feature maps $X^e \in \mathbb{R}^{c \times h \times w}$ enhanced by the causal context pool.

\textbf{Causal context pool.} We maintain a causal context pool $Q \in \mathbb{R}^{n \times h \times w}$ during the network training phase, where $Q_i$ denotes the context of all $i^{th}$ class images. $Q$ ceaselessly stores all contextual information maps of each class by accumulating the activation map of the highest probability class. Then, it projects all contexts of each class as attentions onto the feature maps of the last convolutional layer to produce enhanced feature maps. The idea behind using a causal context pool is not only to cut off the negative effect of entangled context on image feature maps but also to spotlight the positive region of the image feature maps for boosting localization performance.

\textbf{Combinational part.} The input of the combinational part is class activation maps $M$ generated from the CAM module, and the corresponding output is a localization map $H \in \mathbb{R}^{h \times w}$ calculated by Eq.~\eqref{eq:ccam}: First, the combinational part ranks the activation maps from the highest probability class to the lowest. Second, it combines these sorted activation maps by a combinational function as Eq.~\eqref{eq:ccam}.

With all the key modules presented above, we would give a brief illustration of the data flow in our network. Given an image $I$, we first forward $I$ to the backbone to produce feature maps $X$. $X$ is then fed into the following two parallel CAM branches. The first CAM branch produces initial prediction scores $S$ and class activation maps $M$. Then, the causal context pool $Q$ would be updated by fusing the activation map of the highest probability class in $M$ as follows:
\begin{equation}
   \begin{aligned}
      Q_\pi = BN(Q_\pi + \lambda \times BN(M_\pi))),
   \end{aligned}
   \label{eq:update_causal_context_pool}
\end{equation}
where $\pi = argmax(\{s_1, s_2, \ldots, s_n\})$, $\lambda$ denotes the update rate, and $BN$ denotes the batch normalization operation. The second branch is responsible for producing more accurate prediction scores $S^e$ and class activation maps $M^e$. The input of the second branch is enhanced feature maps $X^e$ projected by the context among causal context pool $Q$ of the highest probability class generated from the first branch. More concretely, the feature enhancement can be calculated as
\begin{equation}
   \begin{aligned}
      X^e = X + X\odot Conv_{1 \times 1}(Q_\pi),
   \end{aligned}
\end{equation}
where $\odot$ denotes matrix dot product. In the combinational part, we first build a localization map $H \in \mathbb{R}^{h \times w}$ by aggregating all activation maps from the highest to the lowest probability class using a specific combinational function~\cite{yang2020combinational} in Eq.~\eqref{eq:ccam}. Then, we use the simple thresholding technique proposed by~\cite{zhou2016learning} to generate a bounding box $B$ from the localization map. Finally, the bounding box $B$ and prediction scores $S^e$ as the final prediction of CI-CAM.

\subsection{Training Objective}
\label{sec:training_objective}
During the phase of training, our proposed network learns to minimize image classification losses for both classification branches. Given an image $I$, we can obtain initial prediction scores $S=\{s_1, s_2, \ldots, s_n\}$ and more accurate prediction scores $S^e=\{s_1^e, s_2^e, \ldots, s_n^e\}$ of the two classifiers shown in Figure~\ref{network_architecture_graph}. We follow a naive scheme to train the two classifiers together in an end-to-end pipeline using the following loss function $L$.

\begin{equation}
   \begin{aligned}
      L = \left(-\sum_{i=1}^n s_i^{\ast} log(s_i)\right) + \left( -\sum_{i=1}^n s_i^{\ast} log(s_i^{e})\right),
   \end{aligned}
\end{equation}
where $s^{\ast}$ is the ground-truth label of an image.

\section{Experiments}

\subsection{Datasets and Evaluation Metrics}
\textbf{Datasets.} The proposed CI-CAM was evaluated on two public datasets, \ie, \textbf{CUB-200-2011}~\cite{wah2011caltech} and \textbf{ILSVRC 2016}~\cite{russakovsky2015imagenet}. 1) \textbf{CUB-200-2011} is an extended version of Caltech-UCSD Birds 200 (CUB-200)~\cite{welinder2010caltech} containing $200$ bird species which focuses on the study of subordinate categorization. Based on the CUB-200, CUB-200-2011 adds more images for each category and labels new part localization annotations. CUB-200-2011 contains $5,994$ images in the training set and $5,794$ images in the test set. Each image of CUB-200-2011 is annotated by the bounding boxes, part locations, and attribute labels. 2) \textbf{ILSVRC 2016} is the dataset originally prepared for the ImageNet Large Scale Visual Recognition Challenge (ILSVRC). It contains $1.2$ million images of $1,000$ categories in the training set, $50,000$ in the validation set and $100,000$ images in the test set. For both datasets, we only utilize the image-level classification labels for training, as constrained by the problem setting in WSOL.

\textbf{Evaluation Metrics.} We leverage the \textbf{classification accuracy (cls) and localization accuracy (loc)} as the evaluation metrics for WSOL. The former includes Top-1 and Top-5 classification accuracy, while the latter includes Top-1, Top-5, and GT-know localization accuracy. Top-1 classification accuracy denotes the accuracy of the highest prediction score (likewise for localization accuracy). Top-5 classification accuracy denotes that if one of the five predictions with the highest score is correct, it counts as correct (likewise for localization accuracy). GT-know localization accuracy is the accuracy that only considers localization regardless of classification result compared to Top-1 localization accuracy~\cite{mai2020erasing}.

\subsection{Implementation Details}
We use PyTorch and PaddlePaddle for implementation, both achieve similar performance. We adopt the VGG16~\cite{simonyan2014very} pre-trained on ImageNet~\cite{russakovsky2015imagenet} as the backbone. We insert four non-local blocks to the backbone before every bottleneck layer excluding the first one. The newly added blocks are randomly initialized except for the batch normalization layers in the non-local blocks, which are initialized as zero. We use Adam~\cite{kingma2014adam} to optimize CI-CAM with $\beta_1 = 0.9$, $\beta_2 = 0.99$. We finetune our network with the learning rate $0.0005$, batch size $6$, update rate $\lambda$ $0.01$, and epoch $100$ on the CUB-200-2011 (and learning rate $0.0001$, batch size $40$, update rate $\lambda$ $0.01$, and epoch $20$ on the ILSVRC 2016). At test stage, we resize images to $224 \times 224$. For object localization, we produce the localization map with the threshold $\theta=0.0$, followed by segmenting it to generate a bounding box. The source code will be made public available.

\begin{table}[t]
   \centering
   \caption{Performance on the CUB-200-2011 test set. $\ast$ indicates our re-implemented results.}
   
   \label{sort_cub}
   \begin{tabular}{lcc}
   \toprule
   Methods          & Top-1 Cls(\%) & Top-1 Loc(\%) \\
   \midrule
   VGG-CAM~\cite{zhou2016learning}             &  71.24        &   44.15       \\
   VGG-ACoL~\cite{zhang2018adversarial}             &  71.90         &    45.92      \\
   VGG-ADL~\cite{choe2019attention}              &  65.27         &     52.36     \\
   VGG-DANet~\cite{xue2019danet}      &  75.40     &   52.52      \\ 
   VGG-NL-CCAM~\cite{yang2020combinational}         &  73.4        &    52.4      \\ 
   VGG-MEIL~\cite{mai2020erasing}             &  74.77       &    57.46      \\
   VGG-Rethinking-CAM~\cite{bae2020rethinking}   &  74.91        &     \textbf{61.3}     \\
   \hline
   VGG-Baseline$^\ast$~\cite{yang2020combinational}          &  74.45        &     52.54     \\
   VGG-CI-CAM(ours)     &   \textbf{75.56} ( 1.11 $\uparrow$ )      &   58.39 ( 5.85 $\uparrow$)      \\
   \bottomrule
   \end{tabular}
\end{table}

\begin{table}[t]
   \centering
   \caption{Performance on the ILSVRC 2016 validation set. $\ast$ indicates the our re-implemented results.}
   \label{sort_imagenet}
   \begin{tabular}{lcc}
   \toprule
   Methods          & Top-1 Cls(\%) & Top-1 Loc(\%) \\
   \midrule
   VGG-CAM~\cite{zhou2016learning}               &  66.6    &     42.8     \\
   VGG-ACoL~\cite{zhang2018adversarial}             &  67.5       &  45.83        \\
   VGG-ADL~\cite{choe2019attention}              &  69.48        &    44.92      \\
   VGG-NL-CCAM~\cite{yang2020combinational}          &  72.3        &     \textbf{50.17}     \\ 
   VGG-MEIL~\cite{mai2020erasing}             &  70.27        &     46.81     \\
   VGG-Rethinking-CAM~\cite{bae2020rethinking}   &  67.22        &      45.4    \\
   \hline
   VGG-Baseline$^\ast$~\cite{yang2020combinational}          &  72.15       &    48.55      \\
   VGG-CI-CAM(ours)     & \textbf{72.62} ( 0.47 $\uparrow$ )       &   48.71   ( 0.16 $\uparrow$ )    \\
   \bottomrule  
   \end{tabular}
\end{table}

\subsection{Comparison with State-of-the-Art Methods}

We compare CI-CAM with other state-of-the-art methods on the CUB-200-2011 and ILSVRC 2016 shown in Table~\ref{sort_cub} and Table~\ref{sort_imagenet}.

on the CUB-200-2011, we observe that our CI-CAM significantly outperforms the baseline and can be on par with the existing methods under all the evaluation metrics: CI-CAM yields Top-1 classification accuracy of $75.56\%$, which is $1.11\%$ higher than the baseline and brings about $5.85\%$ improvement to the baseline on Top-1 localization accuracy. Compared with the current state-of-the-art method in terms of classification accuracy, \ie, DANet~\cite{xue2019danet}, CI-CAM outperforms it by $0.16\%$. While under the localization metric, CI-CAM brings a significant performance gain of $5.87\%$ over DANet. Compared with the current state-of-the-art method in terms of Top-1 localization accuracy, \ie, Rethinking-CAM, our model yields a slightly lower localization accuracy but outperforms it on classification. In conclusion, the introduction of causal inference to WSOL task is effective for on both object localization and classification.


For more general scenarios as in ILSVRC 2016 which suffer less from the ``entangled context'' due to the huge amount of images and various backgrounds, CI-CAM can also perform on par with the state of the arts. Compared with the NL-CCAM model (Baseline$\ast$ re-implemented by ourselves), which enjoys state-of-the-art classification and localization accuracy simultaneously, CI-CAM yields a slightly higher result under both metrics. Compared with  MEIL~\cite{xue2019danet}, CI-CAM brings a significant performance gain of $2.35\%$ and $1.9\%$ in classification and localization accuracy over it, respectively.   

\begin{table*}[t]
   \centering
   \caption{Ablation study on the CUB-200-2011 and ILSVRC 2016 datasets. 1) TwoC: Two classifiers, 2)ConPool: Causal context pool. }
   \label{ablation_study_network}
   \begin{tabular}{c|cc|ccccc}
   \toprule
   Dataset & TwoC & ConPool & Top-1 Cls(\%) & Top-5 Cls(\%) & Top-1 Loc(\%) & Top-5 Loc(\%) & GT-know Loc(\%) \\ 
   \midrule
   \multirow{2}*{\tabincell{c}{CUB-200-2011 \\ test set}}
   &$\surd$  & &74.43&91.85&57.16&70.12&75.37 \\
   ~&$\surd$  &$\surd$  &\textbf{75.56}& \textbf{92.03}&\textbf{58.39}& \textbf{70.54}&\textbf{75.68} \\
   \hline
   \multirow{2}*{\tabincell{c}{ILSVRC 2016 \\ val set}}
   
   &$\surd$  &&72.60&90.90&48.25&58.20&61.82\\
   ~&$\surd$  &$\surd$  &\textbf{72.62}&\textbf{90.93}&\textbf{48.71}&\textbf{58.76}&\textbf{62.36} \\
   \bottomrule
   \end{tabular}
\end{table*}

\subsection{Ablation Study}

To better understand the effectiveness of the causal context pool module, we conducted several ablation studies on the CUB-200-2011 and ILSVRC 2016 using VGG16. The results of our ablation studies are illustrated in Table~\ref{ablation_study_network}.

Comparing the first-row and the second-row experimental results on the CUB-200-2011 dataset, employing a causal context pool can comprehensively improve the accuracy of classification and localization. Especially it has increased by $1.13\%$ and $1.23\%$ in the Top-1 classification accuracy and the Top-1 localization accuracy, respectively. Meanwhile, it also improves by $0.18\%$, $0.42\%$, and $0.31\%$ in the Top-5 classification accuracy, Top-5 localization accuracy, and GT-know localization accuracy, respectively. In addition, comparing the third-row and the fourth-row experimental results on the ILSVRC 2016 dataset, we are surprised to find that the causal context pool also performs well on the ILSVRC 2016 dataset which suffers less from the ``entangled context'' due to the huge amount of images and various backgrounds. More specifically, using a causal context pool has improved by $0.46\%$, $0.56\%$, and $0.54\%$ in the Top-1, Top-5, and GT-know localization accuracy. At the same time, the Top-1 and Top-5 classification accuracy has been increased by $0.02\%$ and $0.03\%$ by using a causal context pool, respectively. 

In conclusion, employing a causal context pool can improve classification and localization together on the CUB-200-2011 dataset. And the main improvement of employing the causal context pool on the ILSVRC 2016 dataset is localization accuracy. The experimental results from Table~\ref{ablation_study_network} verify that the introduced causal context pool module can boost the accuracy in the WSOL task.

\begin{table}[t]
   \centering
   \caption{Ablation studies of casual context pool update rate $\lambda$ on the CUB-200-2011 test set (threshold $\theta=0.0$).}
   \label{ablation_study_update_rate}
   \begin{tabular}{lccc}
   \toprule
   Update rate $\lambda$  & Top-1 Cls(\%) & Top-1 Loc(\%) & GT-know Loc\\
   \midrule

   0.001 & 74.28& 58.28& 76.37 \\
   0.002 & 74.77& \textbf{59.23}& \textbf{77.80} \\
   0.005 & 74.49& 58.27& 76.94 \\
   0.01 & \textbf{75.56}& 58.39& 75.68 \\
   0.02 & 74.58& 57.90& 76.54 \\
   0.04 & 74.53& 59.03& 77.63 \\
   0.08 & 74.20& 58.70& 77.77 \\
   \bottomrule
   \end{tabular}
\end{table}

\subsection{Analysis}

As shown in the Eq.~\eqref{eq:update_causal_context_pool}, we introduce a hyperparameter $\lambda$ for updating the causal context pool. Besides, we use a necessary hyperparameter segmentation threshold $\theta$ for segmenting bounding box. Therefore, we will discuss their effects on detection performance when $\lambda$ and $\theta$ take different values in this section on the CUB-200-2011 dataset.

1) \textit{Update rate $\lambda$}. To inspect the effect of the update rate $\lambda$ on classification and localization accuracy, we report the results of using different values of $\lambda$ shown in Table~\ref{ablation_study_update_rate}. By comparing the results we can observe that the update rate $\lambda$ has a great impact on the classification and localization accuracy of the model, especially in GT-know localization ($75.68\%$ \emph{vs} $77.80\%$). The highest Top-1 classification accuracy outperforms the lowest Top-1 classification accuracy by $1.36\%$, and the highest Top-1 localization accuracy outperforms the lowest Top-1 localization accuracy by $1.33\%$. However, in these experiments, there is no $\lambda$ that performs best in both classification and localization. Therefore, when we choose the update rate $\lambda$, we should determine it according to the specific needs of the task.

\begin{figure}[t]
   \centering
   \includegraphics[width=0.8\linewidth]{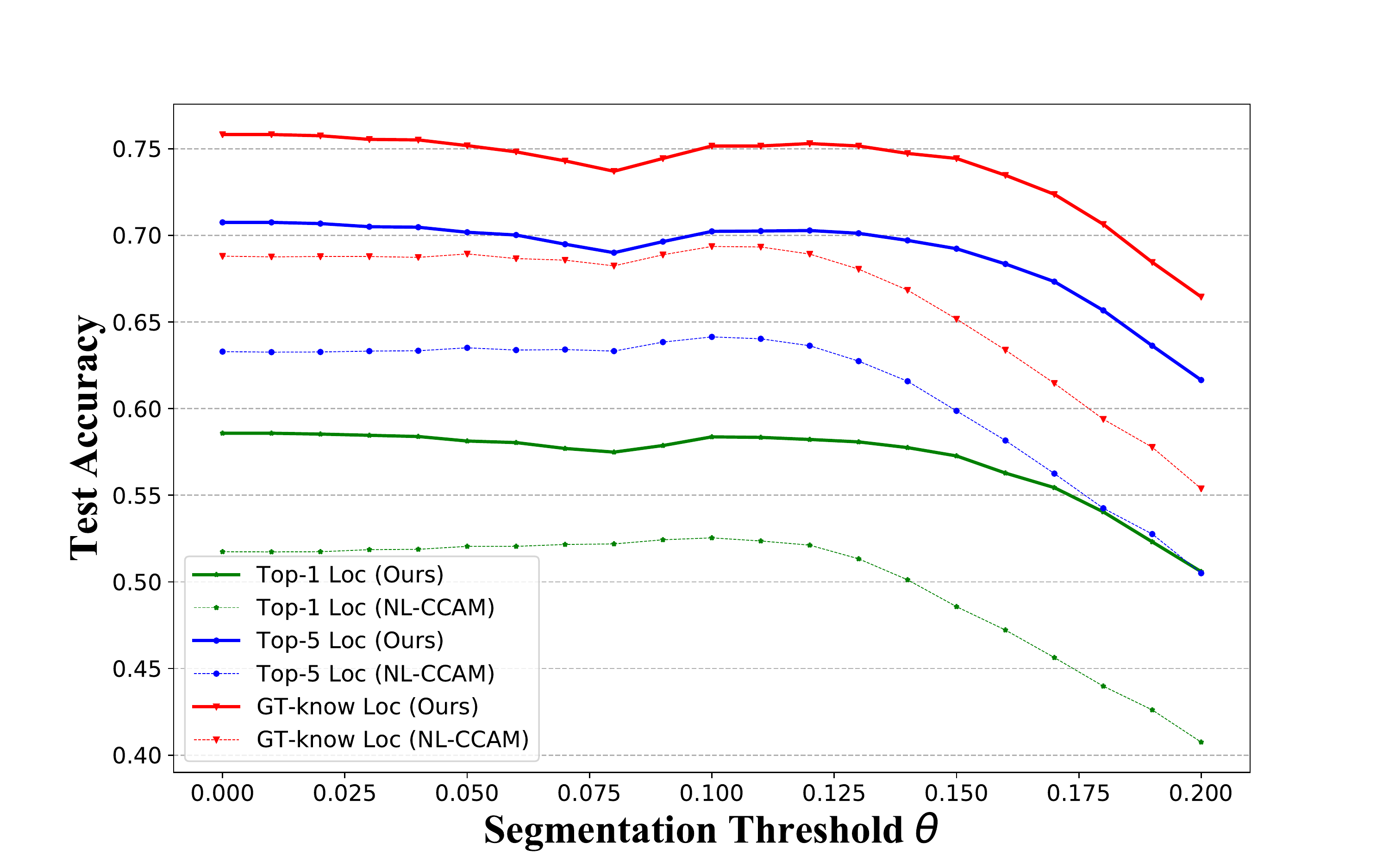}
   \caption{Ablation studies of threshold $\theta$ on the CUB-200-2011 test set (casual context pool update rate $\lambda=0.01$). }
   \label{ablation_study_seg_thr_graph}
 \end{figure}

2) \textit{Segmentation threshold $\theta$}. Although $\theta$ does not participate in the training of the model, it still plays a very important role in object localization. If the value of $\theta$ is low, the detector tends to introduce some highlighted background area around the object. So previous methods~\cite{zhou2016learning, zhang2018adversarial} used segmentation threshold $\theta=0.2$, and NL-CCAM used segmentation threshold $\theta=0.12$. Based on a large segmentation threshold, they tend to filter out low-light object regions and focus on the most discriminative part of the object rather than the whole object. Fortunately, we leverage a causal context pool to resolve the above problem by making the boundary of object and co-occurrence context clearer. 

\begin{figure*}[t]
   \centering
   \includegraphics[width=0.84\linewidth]{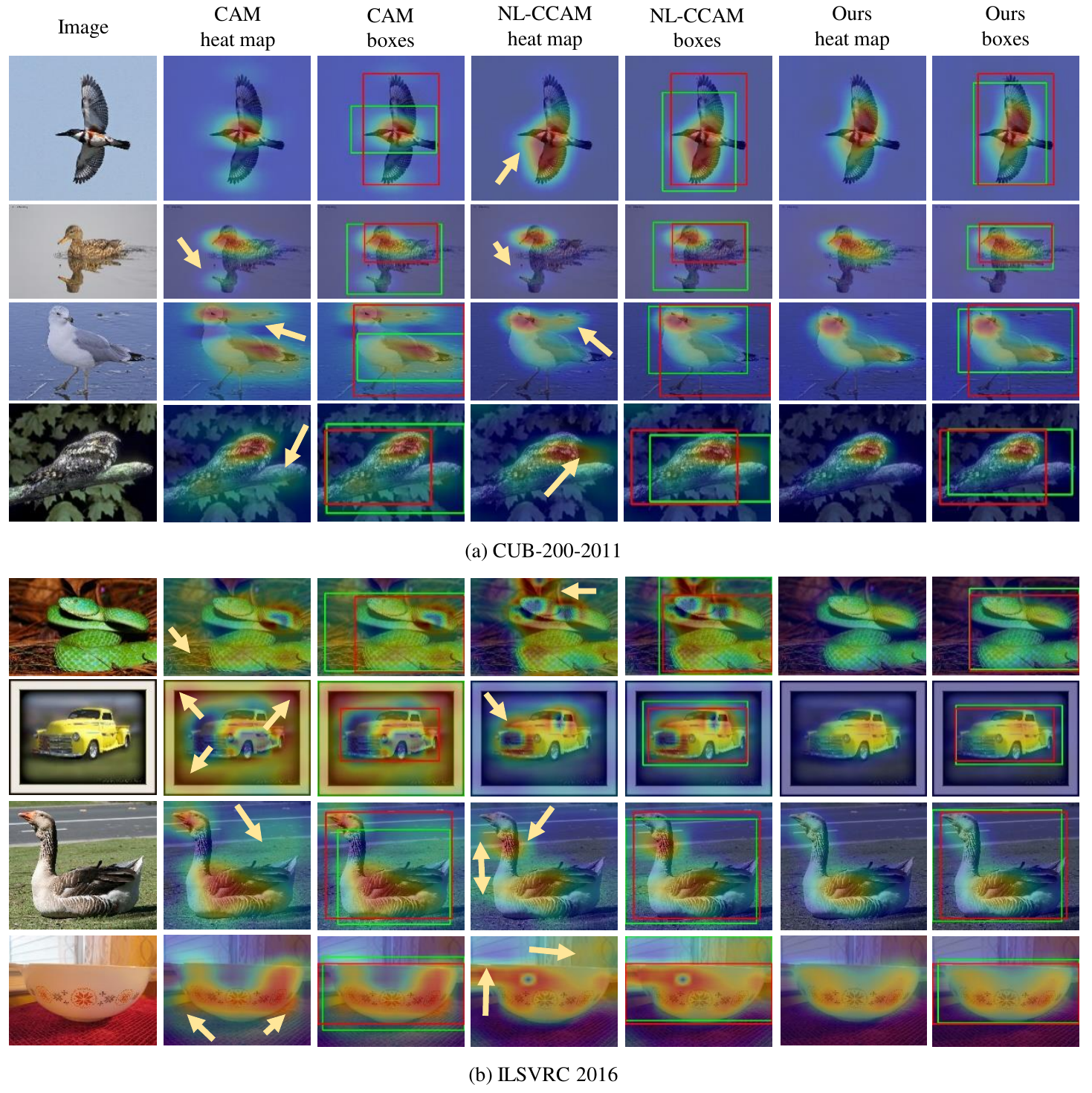}
   \caption{Qualitative object localization results compared with the CAM and NL-CCAM methods. The predicted bounding boxes are in green, and the ground-truth boxes are in red. The yellow arrows indicate the regions suffer from entangled context.}
   \label{result_images}
   \vspace{0.5cm}
\end{figure*}

To inspect the effect of causal context pool on classification and localization accuracy, we test the different segmentation threshold $\theta$ shown in Figure~\ref{ablation_study_seg_thr_graph}. Firstly, we find that the best localization of NL-CCAM is $\theta = 0.1$. When $\theta$ becomes smaller or larger, it will reduce the localization accuracy. Secondly, we can observe that the localization accuracy of CI-CAM is higher than that of NL-CCAM. Especially we obtain the highest Top-1 localization, Top-5 localization, and GT-know localization accuracy when $\theta = 0.0$, which means that CI-CAM can locate a larger part of the object without including the background. Therefore, we can indirectly conclude that CI-CAM is better in dealing with the boundary between the instance and the co-occurrence context background. To illustrate the effect of CI-CAM more vividly, we present the localization maps of the CAM, NL-CCAM, and our model on the CUB-200-2011 as well as the ILSVRC 2016 datasets in Figure~\ref{result_images}. Our visualization in Figure~\ref{result_images} indicates our method is effective in dealing with the co-occurrence context.

\section{Conclusions}
In this paper, we targeted the ``entangled context'' problem in the WSOL task, which remains unnoticed and unexplored by existing efforts. Through analyzing the causal relationship between image features, context, and image labels using a structural causal model, we pinpointed the context as a confounder and tried to utilize probability formula transformation to cut off the link between context and image features. Based on the causal analysis, we proposed an end-to-end CI-CAM model, which uses a causal context pool to accumulate all contexts of each class, and then re-project the fused contexts to the feature maps of convolutional layers to make the feature boundary clearer. To our knowledge, we have made a very early attempt to apprehend and approach the ``entangled context'' issue for WSOL. Extensive experiments have demonstrated that the ``entangled context'' is a practical issue within the WSOL task and our proposed method is effective towards it: CI-CAM achieved the new state-of-the-art performance on the CUB-200-2011 and performed on par with the state of the arts. 


\section{Acknowledge}
This work was supported by the National Natural Science Foundation of China (U19B2043, 61976185), Zhejiang Natural Science Foundation (LR19F020002), Zhejiang Innovation Foundation(2019R52002), CCF-Baidu Open Fund under Grant No. CCF-BAIDUOF2020016, and the Fundamental Research Funds for the Central Universities.

\bibliographystyle{ACM-Reference-Format}
\balance
\bibliography{main}

\end{document}